\documentclass[pdflatex,sn-mathphys,Numbered]{sn-jnl}

\usepackage{graphicx}%
\usepackage{multirow}%
\usepackage{amsmath,amssymb,amsfonts}%
\usepackage{amsthm}%
\usepackage{mathrsfs}%
\usepackage[title]{appendix}%
\usepackage{colortbl}
\usepackage{xcolor}%
\usepackage{textcomp}%
\usepackage{manyfoot}%
\usepackage{booktabs}%
\usepackage{algorithm}%
\usepackage{algorithmicx}%
\usepackage{algpseudocode}%
\usepackage{listings}%
\usepackage{tabularx}
\usepackage{placeins}
\usepackage{multirow}

\def\doi#1{\href{https://doi.org/\detokenize{#1}}{\url{https://doi.org/\detokenize{#1}}}}

\begin{document}

\title[Training a segmentation model with complementary datasets]{One model to use them all: Training a segmentation model with complementary datasets}


\author*[1]{\fnm{Alexander C.} \sur{Jenke}}\email{alexander.jenke@nct-dresden.de}
\author[1,3]{\fnm{Sebastian} \sur{Bodenstedt}}
\author[2,3,4]{\fnm{Fiona R.} \sur{Kolbinger}}
\author[2,3]{\fnm{Marius} \sur{Distler}}
\author[2,3]{\fnm{Jürgen} \sur{Weitz}}
\author[1,3]{\fnm{Stefanie} \sur{Speidel}}

\affil*[1]{\orgdiv{Department of Translational Surgical Oncology}, 
\orgname{National Center for Tumor Diseases (NCT/UCC) Dresden}, 
\orgaddress{\street{Fetscherstraße 74}, \city{Dresden}, \country{Germany}}; German
Cancer Research Center (DKFZ), Heidelberg, Germany; Faculty of Medicine and
University Hospital Carl Gustav Carus, Technical University Dresden, Dresden,
Germany; Helmholtz-Zentrum Dresden-Rossendorf (HZDR), Dresden}
\affil[2]{\orgdiv{Department of Visceral, Thoracic and Vascular Surgery}, \orgname{University Hospital and Faculty of Medicine Carl Gustav Carus, Technical University Dresden}, \orgaddress{\street{Fetscherstraße 74}, \city{Dresden}, \state{Saxony}, \country{Germany}}}
\affil[3]{\orgdiv{Centre for Tactile Internet with Human-in-the-Loop (CeTI)}, \orgname{Technical University Dresden}, \orgaddress{\street{CeTI Exzellenz-Cluster
}, \city{Dresden}, \state{Saxony}, \country{Germany}}}
\affil[4]{\orgdiv{Weldon School of Biomedical Engineering}, \orgname{Purdue University}, \orgaddress{\city{West Lafayette}, \state{Indiana}, \country{USA}}}

\abstract{
\textit{Purpose}
Understanding a surgical scene is crucial for computer-assisted surgery systems to provide any intelligent assistance functionality. 
One way of achieving this so-called scene understanding is via scene segmentation, where a method classifies every pixel of a frame and therefore identifies the visible structures and tissues. 
Recently, progress on fully segmenting surgical scenes has been made using machine learning (ML). 
However, such ML models require large amounts of annotated training data, containing examples of all relevant object classes. 
Such fully annotated datasets are hard to create, as every pixel in a frame needs to be annotated by medical experts and, therefore, are rarely available. 
In this work, we therefore propose a method to combine multiple partially annotated datasets, which provide complementary annotations, into one model, enabling better scene segmentation and the use of multiple readily available datasets.

\textit{Methods} 
Our method aims to combine available data with complementary labels by leveraging mutual exclusive properties to maximize information.
Specifically, we propose to use positive annotations of other classes as negative samples and to exclude background pixels of binary annotations, as we cannot tell if they contain a class not annotated in the frame but predicted by the model.

\textit{Results}
We evaluate our method by training a DeepLabV3 model on the publicly available Dresden Surgical Anatomy Dataset, which provides multiple subsets of binary segmented anatomical structures. 
Our approach successfully combines 6 classes into one model, increasing the overall Dice Score by 4.4\% compared to an ensemble of models trained on the classes individually. By including information on multiple classes, we were able to reduce the confusion between stomach and colon by 24\%.

\textit{Conclusion} 
Our results demonstrate the feasibility of training a model on multiple complementary datasets.
This paves the way for future work further alleviating the need for one large, fully segmented datasets.
}


\maketitle

\section{Introduction}\label{sec:introduction}
The understanding of the visible surgical scene is key for computer-assisted surgery (CAS) systems to understand the current situation and provide adapted assistance functions.
One approach of achieving this, is via the use of full scene semantic segmentation models, which are able to classify every visible part of the scene.
These models provide the basis for recognizing the current situation or actions, and enable useful assistance functions. 

In recent work, progress has been made in improving surgical semantic scene segmentation by the use of temporal context \cite{LSTM_SSSS}, stereo vision \cite{StreoScenNet}, simulated data \cite{SimSSS}, or weak label annotations \cite{easylabels}.
Further steps  towards solving this task in the surgical setting have been taken by the Robotic Scene Segmentation Challenge \cite{RSS}, and the HeiSurF Challenge \cite{HeiSurF}.
Both EndoVis\footnote{\url{http://endovis.org}} sub-challenges provided annotations of 11 and 21 classes, respectively, including surgical tools and human anatomy, and challenged participants to semantically segment all of them.

Nevertheless, a major bottleneck for clinical translation of surgical data science (SDS) applications remains the availability of such datasets \cite{MaierHein22SDS}, due to the high amount of time required for experts, to create such segmentation annotations.
This is even more challenging for full scene semantic segmentation, which requires a pixel-wise annotation of multiple classes for the complete frame.

This issue was sidestepped in the recently published Dresden Surgical Anatomy Dataset (DSAD) \cite{DSAD} by simply providing binary segmentations.
The dataset contains 11 classes of anatomical structures, split into multiple subsets providing one class each.
This way, the authors were able to publish over 13000 expert-approved semantic segmentations.

At present there are a handful of datasets providing different annotations relevant to the field of SDS \cite{MaierHein22SDS}. 
These datasets can differ in the granularity of classes and the classes annotated in general, depending on the protocol used during annotation.
Recent works have presented different methods that rely only on partial annotations for training a CT segmentation model, improving the usage of existing knowledge \cite{shi2021, MultiTalent, dmitriev2019learning, yan2020learning}.
The used approaches range from dataset specific backbones and pseudo label generation \cite{yan2020learning}, over merging unlabeled classes with the background class and adding a mutual exclusion constraint \cite{shi2021}, to simply masking unlabeled classes during loss calculation \cite{MultiTalent}.

We assume that learning to segment multiple classes causes a single model to develop a better understanding of the scene, leading to better segmentation performance. 
We therefore propose the usage of information from mutual-exclusivity, which can be applied on top of any state-of-the-art approach.
In this work we apply it in addition to the masking during loss calcation \cite{MultiTalent}.
We demonstrate the feasibility of training a model for multi-class organ segmentation of laparoscopic surgery images on multiple complementary dataset, thereby overcoming the data bottleneck challenge in SDS.
The code and models are publicly available on \url{https://gitlab.com/nct\_tso\_public/dsad-segmentation/}.

\section{Methods}\label{sec:methods}
\begin{figure}[t!]
    \centering
    \includegraphics[width=.95\textwidth]{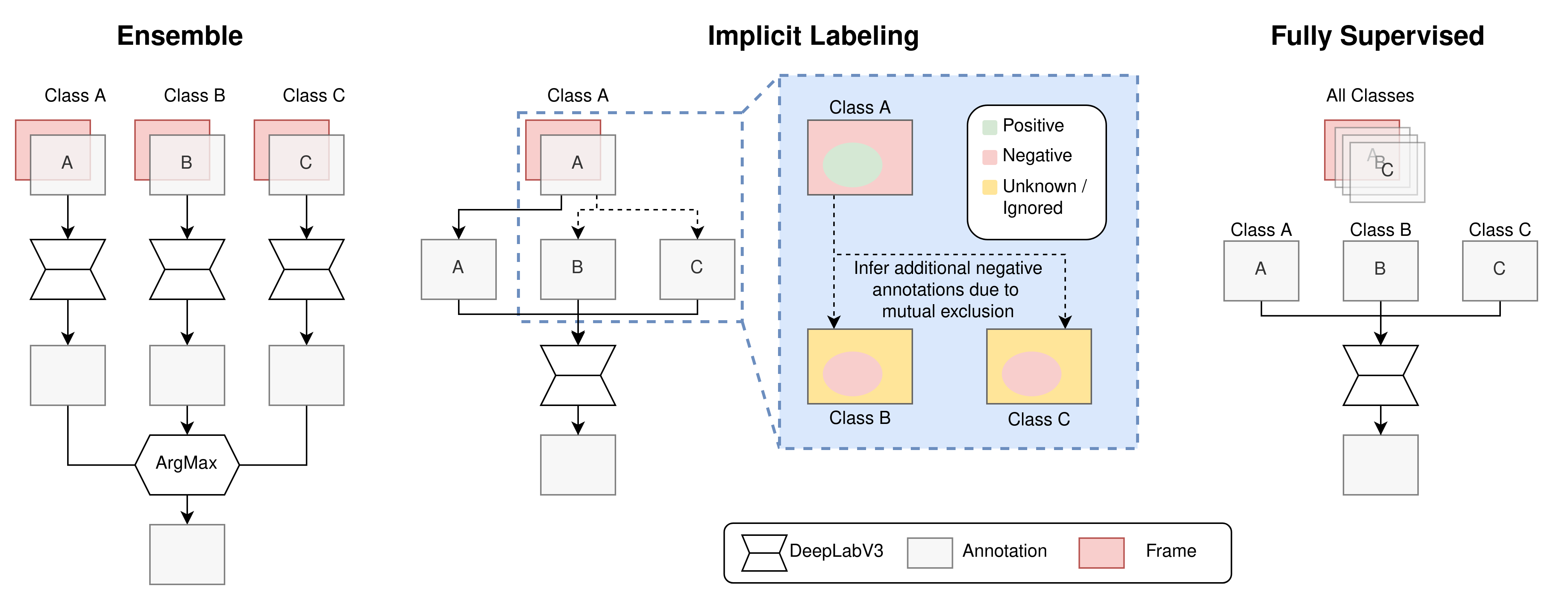}
    \caption{A flowchart describing the flow of annotation data through the three architectures used in this work.
    The proposed implicit label generation is shown in the blue box.}
    \label{fig:flowchart}
\end{figure}

In this section, we introduce our proposed method and define an upper and lower baseline to compare the performance of the proposed method to.
A visualization of the architectures is shown in Figure \ref{fig:flowchart}.
Further, we provide a definition of how the average dice scores are calculated in this work.

\subsection{Baselines}
A naive approach to combine multiple datasets is to train one model per dataset, and subsequently building an ensemble to merge their predictions into one final multi-class prediction.
In case of one binary segmenting model per class, this kind of ensemble prediction can be achieved by applying the argmax over the different sigmoid outputs per pixel. 
This is followed by a threshold to determine if the most likely class is predicted positive, otherwise the background class is assigned. 
In this work this approach will be used as the lower baseline and is referred to as ensemble (EN).

Alternatively a single model can be trained on a single fully labeled dataset. 
As this requires a large fully labeled dataset, which is hard to obtain but provides the maximum amount of information, this approach will be used as the upper baseline in this work and is referred to as fully supervised (FS).

\subsection{Implication based labeling}
To overcome the ambiguity inherent to merging multiple models and lost information between classes, while sidestepping the need for a single large fully annotated dataset, we propose to combine the classes provided by multiple datasets into one model. 
This reduces computational costs during inference as only one output from one model is required and lowers the requirements we pose against the dataset, as not all classes need to be included in every dataset.
In addition, the model could benefit from the shared knowledge among the different classes.

The model takes single images of the datasets as input and outputs a class probability for every pixel and every class.
For every pixel, the final class is selected based on the highest predicted class probability.
The outputs are normalized by a sigmoid and therefore each value is independent of the outputs of other classes.
On one hand, this setup does not prevent the model from predicting multiple classes per pixel, instead this has to be enforced by the loss during training. 
On the other hand, this setup allows us to access the predicted probability of every class separately, which is required for dealing with the problem of training from incomplete knowledge in the datasets, as each dataset provides only information on its contained class and not the classes introduced by other datasets.

Assuming every pixel in the target semantic scene segmentation problem is exactly part of one class (mutually exclusivity of classes), this can be used to maximize the information provided by a partially annotated dataset by applying the following rules:
\begin{enumerate}
    \item A positive annotation of one class implies a negative annotation of all other classes.
    \item A negative annotation of a class (in a binary annotation the background) provides no information if this region contains other classes. As no implication to other classes can be made, the annotation stays \textit{unknown}.
\end{enumerate}
\noindent The application of these rules is visualized in the blue box of Figure \ref{fig:flowchart}.

In this work we use these implication rules to obtain additional negative samples from datasets not containing the examined class.
Cases in which the annotation stays \textit{unknown} are excluded from the loss calculation by masking, as no decision on the correctness of the prediction can be made.
This method is called implicit labeling (IL) in the following.

The loss per class $c$ can be formulated as: 
\begin{equation}
    L_c = \frac{1}{B} \frac{1}{\hat{P}} \sum_{b,p}{BCE(\hat{y}^{(c)}_{b,p}, y^{(c)}_{b,p}) * \lambda_{b,p,c}}
    \label{eq:merge}
\end{equation}
\begin{equation}
    \lambda_{b,p,c} =
    \begin{cases}
        1, & \text{if } b \text{ is annotated for } c\text{ or }y^{(\hat{c})}_{b,p} = 1, \hat{c} \in C\setminus c\\
        0, & \text{else}\\
    \end{cases}
\end{equation}
where $B$ is the number of images, $b\in[1,B]$, in the batch, and $\hat{P}$ is the number of Pixels, $p\in P$, where $\lambda = 1$, $\hat{y}^{(c)}_{b,p}$ is the prediction of class $c$ for pixel $p$ of image $b$ and $y^{(c)}_{b,p}$ the respective ground truth. 
A pixel $p$ is included in the loss calculation if either the image $b$ is annotated for class $c$ or if $y^{(\hat{c})}_{b,p} = 1$ for any other class $\hat{c}\in C\setminus {c}$, otherwise the pixel is excluded by $\lambda_{b,p,c}$ being set to zero. 
In the second case the ground truth for class $c$ is false, $y^{(c)}_{b,p} = 0$, due to the mutually exclusivity of classes.

\subsection{Metrics}
In this work the model performance is evaluated using the dice score \cite{dice}.
For the average dice score per class, the dice is calculated per image and averaged subsequently over all images. If a class does not occur in either the target nor the prediction, the F1 is not defined, we therefore set the score to one in those cases as the model performed as expected.
The average dice score over all classes is calculated by averaging all average dice scores per class.

Statistical significance is calculated using a two-sided Wilcoxon signed-rank test on the image-wise and class-wise dice scores of our approach against the lower baseline.
The significance is calculated per class. The significance of the mean dice score is calculated by evaluationg all image-wise and class-wise dice scores over all classes.

\section{Evaluation}\label{sec:evaluation}
\subsection{Dataset}
We evaluate our proposed approach using the publicly available DSAD dataset \cite{DSAD}.
This dataset consists of 13,195 laparoscopic images split into 11 subsets with a minimum of 1,000 frames from a minimum of 20 surgeries each. 
In each subset, binary segmentations for one of the 11 classes (abdominal wall, colon, inferior mesenteric artery, intestinal veins, liver, pancreas, small intestine, spleen, stomach, ureter, vesicular glands) are provided.
For the stomach subset, additional masks are available, annotating six of the remaining ten classes (abdominal wall, colon, liver, pancreas, small intestine, spleen) visible in this subset, resulting in one multi-class subset.
This work uses the proposed split \cite{dsad_split} into training, validation, and test set. 
To be able to compare to the multi-class subset, only the binary subsets of the contained classes are used, with the exception of the spleen. 
The spleen class was excluded in this work due to the lack of positive examples in the validation and test split in the multi-class subset.
To examine the ability of IL and EN to join datasets, they were trained by splitting the multi-class subset into six binary sets, one for each class.
The methods trained on these subsets are comparable to the FS approach trained on the combined annotations of the multi-class subset.
As the binary subsets have no overlapping classes, we interpret them as separate datasets in this work, which we aim to join into one model.
All classes fulfill the required mutual exclusiveness. 

\subsection{Trials}
To validate our approach, we conducted five trials to compare our proposed implicit labeling approach to the ensemble approach and the fully supervised approach.
In all experiments, the DeepLabV3 architecture with a ResNet50 backbone \cite{deeplabv3} is used. 
Models are initialized using the default PyTorch pretraining on COCO \cite{COCO}.
All models were trained using PyTorch \cite{pytorch} v1.12 on Ubuntu 18.04 and Nvidia Tesla V100 GPUs, evaluation was done on Ubuntu 20.04 with Nvidia RTX A5000 GPUs.
The images were down scaled to a size of 640x512 pixels for memory and time reasons.
For all trials an initial learning rate of $3\times 10^{-4}$ was used with a scheduler reducing the learn rate by a gamma of 0.9 every 10 epochs.
Further all model were trained for 100 epochs using an AdamW optimizer with weight decay of 0.1 and an cross-entropy loss with a \textit{positive weight} factor for balancing positive and negative pixels. 
The best model per trial was selected according to the dice score on the validation set. 

~\\
The ensemble approach evaluates an ensemble of six models, each trained on one of the classes, serving as the lower baseline.
The prediction of the ensemble was obtained by selecting the class with the highest value over all models per pixel. 
The positive weight for the background was set to 1 and the remaining classes to the negative to positive pixel ratio per class. 
This approach was used in two trials, once trained on the binary subset, and once on the binary sets extracted from the multi-class subset.\\
The fully supervised approach examines the fully supervised method, serving as the upper baseline.
For this a single DeepLabV3 was trained on the multi-class subset.
The positive weight was calculated by the share of positive pixels of the class, negating the share, and normalizing all negated shares with the softmax function.
As this approach requires a fully annotated dataset, this method is only trained once on the multi-class subset.
~\\
In the implicit labeling approach, our proposed method was used to train a single DeepLabV3 on the six binary subsets, which represent the same classes that are available in the multi-class subset. 
The positive weight was set to the negative to positive pixel ratio, while including the positive pixels of other classes as negatives, as described before.
The output of the loss function was masked before aggregation to ignore pixels as required by our approach.
This approach was used in two trials, once trained on the binary subset, and once on the binary sets extracted from the multi-class subset.

\subsection{Results}
The trials were evaluated twice on the up-to-now unseen test split of the dataset, once using the binary subset for each class and once using the multi-class subset. 
As the multi-class subset is based on the stomach subset, the frames and, therefore, the results for this class are identical.

\begin{table}[htbp!]
    \centering
    \caption{Dice score of each class calculated on the testsets of the binary subsets, $\mu$ being the mean over all classes. 
    Every row represents one of the trials, fully supervised (FS), ensemble (EN), and implicit labeling (IL), trained on the binary and multi-class sets.
    The highest value per column and trainset is highlighted in bold.
    The classes Abdominal Wall, Colon, Liver, Pancreas, Small Intestine and Stomach are abreviated by their first three letters.
    Cells are highlited in green if the difference between IL and EN is strongly significant (p$<$.01) or yellow if significant (p$<$.05).}
    \begin{tabular}{c|c|c|c|c|c|c|c|c}
    Trainset & Trial &  Abd & Col & Liv & Pan & Sma & Sto & $\mu$ \\
    \hline
    \multirow{2}{*}{Binary} & IL (ours) & 0.892 & \cellcolor{green!25}\textbf{0.782} & 0.741 & 0.364 & \cellcolor{green!25}\textbf{0.849} & \cellcolor{green!25}\textbf{0.709} & \cellcolor{green!25}\textbf{0.723} \\
     & EN (lower) & \textbf{0.895} & \cellcolor{green!25}0.729 & \textbf{0.745} & \textbf{0.373} & \cellcolor{green!25}0.826 & \cellcolor{green!25}0.503 & \cellcolor{green!25}0.679 \\
    \hline
    \multirow{3}{*}{MultiClass} & FS (upper) & \textbf{0.764} & \textbf{0.451} & \textbf{0.521} & 0.170 & 0.122 & \textbf{0.719} & \textbf{0.458} \\
    & IL (ours) & \cellcolor{yellow!25}0.730 & \cellcolor{green!25}0.381 & \cellcolor{green!25}0.520 & \cellcolor{green!25} \textbf{0.184} & \cellcolor{green!25} \textbf{0.338} & \cellcolor{green!25}0.578 & \cellcolor{green!25}0.445 \\
    & EN (lower) & \cellcolor{yellow!25}0.673 & \cellcolor{green!25}0.320& \cellcolor{green!25}0.490 & \cellcolor{green!25}0.139 & \cellcolor{green!25}0.177 & \cellcolor{green!25}0.699 & \cellcolor{green!25}0.416 \\

    \end{tabular}
    \label{tab:binary}
\end{table}

The results on the binary testset are shown in Table \ref{tab:binary}.
The overall best performance is reached by our proposed IL approach on the binary trainset with an average dice score of 72\% over all classes outperforming EN by 4\%
On the binary trainset all scores of the IL approach are either significantly higher or not significantly lower than the EN approach.
For the trials trained on the multi-class subset the upper baseline, FS, reaches the highest score of 46\%.
For the highest score per class distributes among the FS and IL approach.
On the multi-class trainset all scores of the IL approach are significantly above the EN approach, except for the stomach. 
The performance of all methods is significantly lower for Pancreas and Liver on the multi-class trainset compared to the binary trainset, while IL maintains the highest score.

\begin{table}[htbp!]
    \centering
    \caption{Dice score of each class calculated on the testset of the multi-class subset, $\mu$ being the mean over all classes. 
    Every row represents one of the trials, fully supervised (FS), ensemble (EN), and implicit labeling (IL), trained on the binary and multi-class sets.
    The highest value per column and trainset is highlighted in bold.
    The classes Background, Abdominal Wall, Colon, Liver, Pancreas, Small Intestine and Stomach are abreviated by their first three letters.
    Cells are highlited in green if the difference between IL and EN is strongly significant (p$<$.01) or yellow if significant (p$<$.05).}
    \begin{tabular}{c|c|c|c|c|c|c|c|c|c}
    Trainset & Trial & Bac & Abd & Col & Liv & Pan & Sma & Sto & $\mu$ \\
    \hline
    \multirow{2}{*}{Binary} & IL (ours) & \cellcolor{green!25}0.932 & 0.578 & \cellcolor{green!25}\textbf{0.472} & \cellcolor{green!25}0.222 & \cellcolor{green!25}\textbf{0.503} &  \cellcolor{green!25}\textbf{0.386}  & \cellcolor{green!25}\textbf{0.709} & \cellcolor{green!25}\textbf{0.543} \\
    & EN (lower) & \cellcolor{green!25}\textbf{0.955} & \textbf{0.590} & \cellcolor{green!25}0.363 &\cellcolor{green!25} \textbf{0.375} & \cellcolor{green!25}0.335 & \cellcolor{green!25}0.214 & \cellcolor{green!25}0.503 & \cellcolor{green!25}0.476 \\        
    \hline
    \multirow{3}{*}{MultiClass} & FS (upper)  & \textbf{0.970} & \textbf{0.800} & \textbf{0.450} & \textbf{0.862} & \textbf{0.679} & \textbf{0.913} & \textbf{0.719} & \textbf{0.783} \\
    & IL (ours) &\cellcolor{green!25} 0.905 & \cellcolor{green!25}0.632 &\cellcolor{green!25} 0.251 & \cellcolor{green!25}0.650 & \cellcolor{yellow!25}0.306 & \cellcolor{green!25}0.455 & \cellcolor{green!25}0.578 & \cellcolor{green!25}0.540\\
    & EN (lower) &\cellcolor{green!25} 0.931 &\cellcolor{green!25} 0.728 &\cellcolor{green!25} 0.360& \cellcolor{green!25}0.387 & \cellcolor{yellow!25}0.239 & \cellcolor{green!25}0.304 & \cellcolor{green!25}0.699 & \cellcolor{green!25}0.512 \\
    \end{tabular}
    \label{tab:multi-class}
\end{table}
The results on the multi-class testset are shown in Table \ref{tab:multi-class}.
The overall best performance is reached by the FS approach with an average dice score of 78\%.
On the binary trainset IL outperforms EN by 7\%.
Especially in the classes Colon, Pancreas, Small Intestine and Stomach IL performs significantly better than EN, while EN is better for the Liver.
On the multi-class testset FS has the hightest score on all classes. 
The second highest class is distributed among IL and EN. 
The average performance of IL outperforms EN by 4\%.

\begin{figure}[bp!]
    \centering
    \includegraphics[width=.9\textwidth]{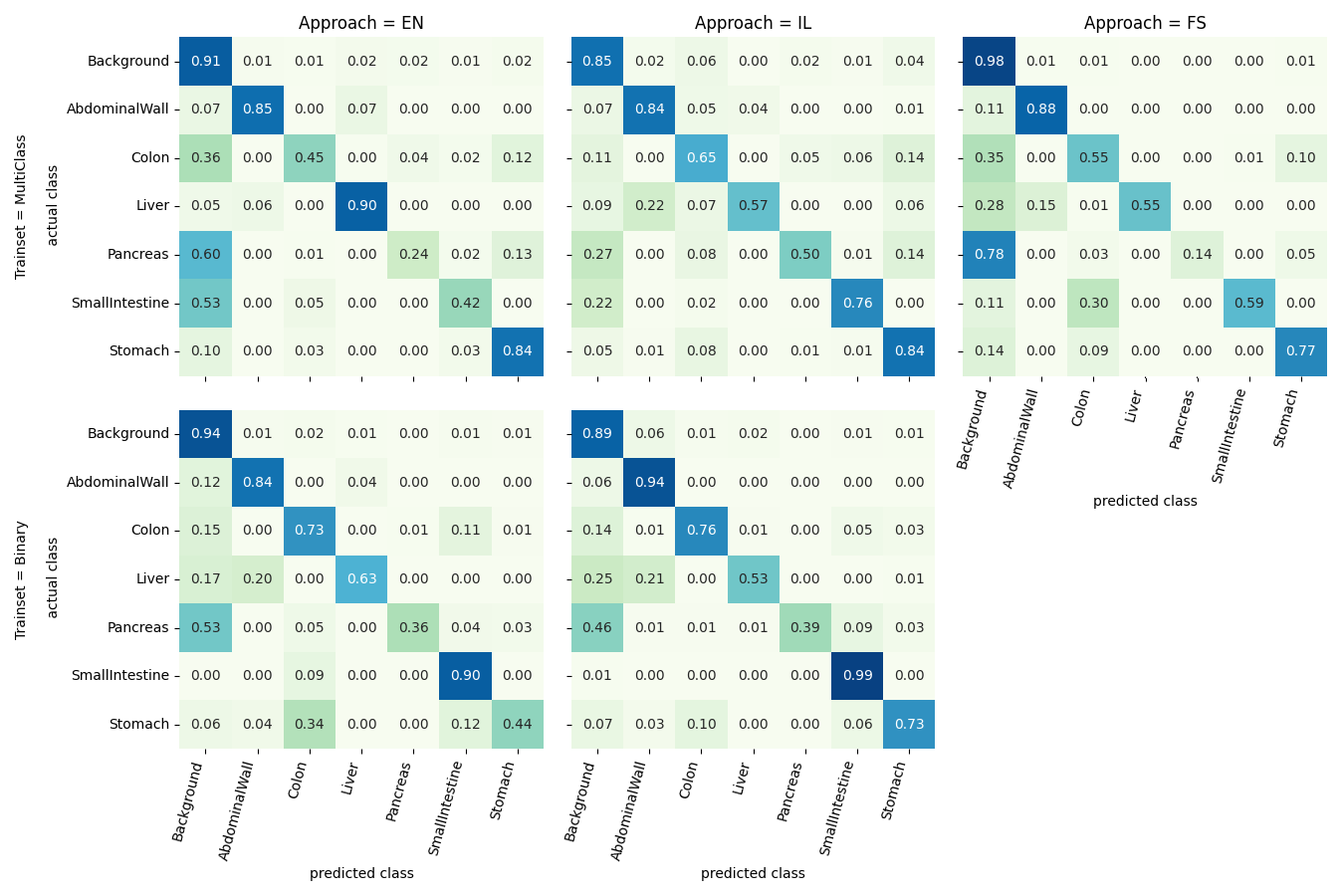}
    \caption{Pixelwise confusion matrix per approach and trainset tested on the multi-class testset.}
    \label{fig:confusion}
\end{figure}

Figure \ref{fig:confusion} shows the pixelwise confusion of the classes for the trials tested on the multi-class testset.
For trials trained on the multi-class trainset IL shows lower confusion with the background for the pancreas and the small intestine than EN.
This means pixels of both classes are less often missed to detect.
Compared to FS, IL less often confuses the small intestine with the colon.
The liver is more often confused with the abdominal wall by IL compared to FS and EN.
For the binary trials, IL lowers the confusion of the stomach with the colon and small intestine.
The confusion of the liver with the abdominal wall and background slightly increases. 

\begin{figure}[htbp!]
    \centering
    \includegraphics[width=.90\textwidth]{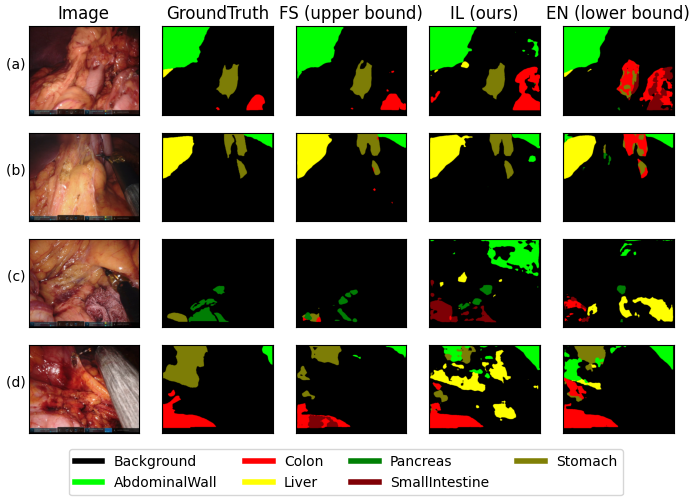}
    \caption{Segmentation results for 3 example images. The columns show the original frame, the ground truth, and the predicted segmentations of the fully supervised (FS), implicit labeling (IL), and ensemble (EN) approaches. Rows (a) and (b) show examples of good IL performance, (c) and (d) show more difficult cases.}
    \label{fig:examples}
\end{figure}
Figure \ref{fig:examples} shows four examples of segmentation results. In the rows (a) and (b), FS and IL are able to segment a single structure with one class while the EN approach is mixing multiple classes.
In rows (c) andd (d) all three approaches produce patched results per organ and detect classes not present.

\begin{table}[htbp!]
    \centering
    \caption{
    Dice score of each class calculated on the testsets of the binary subsets, $\mu$ being the mean over all classes. 
    The first row shows the performance of each separate models in the ensemble before applying the argmax.
    The second row shows a variant of the IL approach if no additional negative samples are inferred.
    The classes Colon, Liver, Pancreas, Small Intestine and Stomach are abbreviated by their first three letters.
    }
    \begin{tabular}{c|c|c|c|c|c|c|c}
    Trial &  Abd & Col & Liv & Pan & Sma & Sto & $\mu$ \\
    \hline
    Separate models of EN         & 0.894 & 0.790 & 0.786 & 0.400 & 0.866 & 0.701 & 0.740 \\
    IL (ours)                     & 0.892 & 0.782 & 0.741 & 0.364 & 0.849 & 0.709 & 0.723 \\
    IL w/o additional neg. labels & 0.897 & 0.766 & 0.783 & 0.326 & 0.832 & 0.633 & 0.706 \\
    \end{tabular}
    \label{tab:ablation}
\end{table}
Table \ref{tab:ablation} shows the results of out ablation studies trained and evaluated on the binary subset.  
The first row shows the dice score of the separate models in the ensemble before applying the argmax.
All classes except the abdominal wall reach higher scores compared to the merged ensemble. 
IL is outperformed on all classes except the stomach. 
The second row shows the results of our IL approach if no additional negative samples are inferred due to mutual exclusion and only the loss masking is applied. 
While this model is better than our proposed IL approach for the classes abdominal wall and liver the remaining classes and the average performance are below IL.

The trials were evaluated with respect to the inference time by inferring 1000 random inputs on an Nvidia RTX A5000. 
The time needed from loading the image to the GPU to downloading the prediction back to the CPU was averaged.
All models were already prepared on the GPU. 
The ensemble approach required $136\pm{}5.8$ ms per frame, the fully supervised and implicit labeling approaches both required $23\pm{}0.1$ ms per frame.
Resulting in 7fps and 43fps, respectively, and therefore a 6x speed up.
Memory consumption on the GPU stayed below 3.2 GB for the ensemble and 2.3 GB for the fully supervised and implicit labeling approaches.

\section{Discussion}\label{sec:discussion}
In summary, this work demonstrates that the implicit labeling approach is able to leverage multiple complementary datasets into one model.
We find that models benefit from a better scene understanding through more learned classes and reach better performance. 
Further, we point out limitations due to the lack of data diversity we came across in the different data subsets.

As Table \ref{tab:ablation} shows the performance on EN merged by argmax drops compared to the performance of its separate models. 
This again proves the importance of being able to train a single model being able to understand and segment all classes. 
Best results are reached by fully annotated datasets, as shown by FS in Tables \ref{tab:binary} and \ref{tab:multi-class}.
But as these are rarely available, our proposed method is a valid alternative in cases where multiple datasets need to be merged.

Tables \ref{tab:binary} and \ref{tab:multi-class} clearly show, that IL significantly outperforms EN in all four combinations of test and trainset of the binary and multi-class subsets.
On the binary train and binary test combination our proposed approach is either within 1\% difference on the dice score or significantly better than the lower baseline.
This supports our initial assumption that the models benefit from better scene understanding due to complementary information.
As shown in Figure \ref{fig:examples} (a) and (b) the knowledge of multiple classes, as in the FS and IL approach, helps the model to select one class.
This missing information in the EN approach leads to segmentations being patched together of multiple classes.
IL is able to solve this issue without the need for fully segmented annotations.

This is demonstrated further in Figure \ref{fig:confusion} by comparing the class confusion of IL to EN.
For all classes except the liver the confusion drops. 
Especially the confusion of colon, small intestine and stomach on the binary trainset, where those classes do look similar in many cases, are reduced by the knowledge of all classes. 
Interestingly the fully supervised model is not able to prevent confusion of the small intestine and the colon, which might be caused by rare occurrence of those classes due to the stomach centric multi-class subset. 
Interestingly, the large increase of negative samples in the IL approach compared to the FS approach does not lead to the model to prefer the background class rather the opposite is the case. 
This might be due to the higher weighting factors of positive pixels used in the loss.
This proves our assumption that models benefit from better scene understanding through more learned classes.

As shown in Table \ref{tab:binary} and \ref{tab:multi-class}, the models perform worse when tested on the subset they are not trained on.
Especially on the Pancreas and Small Intestine the FS approach drops in performance.
This is explained by the fact that the multi-class subset is based on the stomach.
Therefore, all frames contain this structure, limiting the possible viewing angle and distances of other annotated classes.
IL is able to maintain the highest cross-domain performance in both directions.
This cross-domain capabilities can further be seen in the first row of Table \ref{tab:multi-class} where IL is trained on the multi-class trainset and tested on the binary subset without any performance drop on the average dice score compared to the third row which shows the performance of IL tested on the multi-class testset.
EN, in rows two and four, is dropping by 5\% due to the different appearance in the subsets.

The second row of Table \ref{tab:ablation} shows the importance of the implicit labels, as the models performance drops without them. 
This ablation is comparable to the approach presented by Ulrich et al. \cite{MultiTalent}, which further proves the benefit of our method, as by simply adding the implicit knowledge we outperform related work.

When looking at the situation given by the dataset we are limited to a small fully annotated dataset and have more binary data on hand. 
While it is possible to train the FS approach on the small dataset and achieve a good performance when testing on similar data, as shown in Table \ref{tab:multi-class}, we do see the performance drop when the model is applied to different data, as shown in Table \ref{tab:binary}.
The classes Pancreas and Small Intestine drop by 51\% and 79\%, respectively, resulting in dice scores below 20\%.
Considering that the binary dataset is focused around the organ it is tested for, these misclasifications can be considered critical, as the central organ is not recognized.
When training our IL approach, on the other hand, not only more data is available, but also the performance drop in changing data is less severe. 
While we do see dropping performance on classes, if they are no longer in the focus of the frame, no class drops below a dice score of 25\%, as shown in Table \ref{tab:multi-class}.
This proves our IL approach is more applicable in a realistic setting with limited data availability, producing more robust results without the need for complicated preprocessing or much higher computational costs.

Finally, as the inference time results show, the implicit labeling approach combines fast inference with the ability to learn on not fully annotated datasets.
The ensemble is 6 times slower, as it needs to infer 6 models instead of one, setting a limit to the scalability. 
The runtime of the ensemble increases linearly with the number of classes, while our implicit labeling approach allows adding more segmentation data with a negligible effect on runtime.

\section{Conclusion}\label{sec:conclusion}
In this paper, we presented, to the best of our knowledge, the first approach for laparoscopic organ segmentation that combines multiple datasets into one model. 
We accomplished this by applying a combination of masking during loss and mutual exclusion constraints.
We were able to show that segmentation models benefit from a better understanding of the scene in the sense of knowing more classes, improving the overall dice score, reducing confusion between classes and improving generalization to changes in the appearance of the classes.
Further, we were able to show that we do not require all classes to be annotated in a single dataset but rather can combine complementary ones.
The resulting model was able to achieve real-time capable inference speeds of 43fps on an Nvidia RTX A5000 GPU, and additional classes can be added without linearly increasing the runtime.

Even though the already good results, we see potential to further improve the approach.
For example, the model might include weak labels during training in the form of binary presence of classes.
Also, the effects of the use of datasets with overlapping classes should be investigated, as well as the possibility to combine non-complementary datasets, like different levels of detail in the classes.
For example, some datasets only segment instruments in one class, and others split them up into different types.
Further, the use of active learning to selectively annotate missing data might be promising, as it matches very well with the structure of complementary datasets.
Overall, we see great potential in applying our method in multiple settings of semantic segmentation of surgical data science.

\section*{Declarations}
\textbf{Competing interests} The authors declare no conflict of interest.\\
\textbf{Ethical approval} For this type of study, formal consent is not required.\\
\textbf{Informed consent} This article contains patient data from publicly available datasets.\\
\textbf{Funding} This work is funded by the German Federal Ministry of Health (BMG), on the basis of a decision by the German Bundestag, within the "Surgomics" project (Grant Number BMG 2520DAT82),
the German Cancer Research Center (CoBot 2.0),
the German Research Foundation (Deutsche Forschungsgemeinschaft, DFG) as part of Germany’s Excellence Strategy (EXC 2050/1, Project ID 390696704) within the Cluster of Excellence ”Centre for Tactile Internet with Human-in-the-Loop” (CeTI) of the Dresden University of Technology
and by the European Union through NEARDATA under the grant agreement ID 101092644.
FRK is supported by the Joachim Herz Foundation (Add-On Fellowship for Interdisciplinary Life Science).\\
\textbf{Code \& Data availability} The code is available on \url{https://gitlab.com/nct\_tso\_public/dsad-segmentation/}.
The used DSAD dataset is available at \doi{10.6084/m9.figshare.21702600}.

\bibliography{sn-bibliography}


\begin{thebibliography}{17}
\ifx \bisbn   \undefined \def \bisbn  #1{ISBN #1}\fi
\ifx \binits  \undefined \def \binits#1{#1}\fi
\ifx \bauthor  \undefined \def \bauthor#1{#1}\fi
\ifx \batitle  \undefined \def \batitle#1{#1}\fi
\ifx \bjtitle  \undefined \def \bjtitle#1{#1}\fi
\ifx \bvolume  \undefined \def \bvolume#1{\textbf{#1}}\fi
\ifx \byear  \undefined \def \byear#1{#1}\fi
\ifx \bissue  \undefined \def \bissue#1{#1}\fi
\ifx \bfpage  \undefined \def \bfpage#1{#1}\fi
\ifx \blpage  \undefined \def \blpage #1{#1}\fi
\ifx \burl  \undefined \def \burl#1{\textsf{#1}}\fi
\ifx \doiurl  \undefined \def \doiurl#1{\url{https://doi.org/#1}}\fi
\ifx \betal  \undefined \def \betal{\textit{et al.}}\fi
\ifx \binstitute  \undefined \def \binstitute#1{#1}\fi
\ifx \binstitutionaled  \undefined \def \binstitutionaled#1{#1}\fi
\ifx \bctitle  \undefined \def \bctitle#1{#1}\fi
\ifx \beditor  \undefined \def \beditor#1{#1}\fi
\ifx \bpublisher  \undefined \def \bpublisher#1{#1}\fi
\ifx \bbtitle  \undefined \def \bbtitle#1{#1}\fi
\ifx \bedition  \undefined \def \bedition#1{#1}\fi
\ifx \bseriesno  \undefined \def \bseriesno#1{#1}\fi
\ifx \blocation  \undefined \def \blocation#1{#1}\fi
\ifx \bsertitle  \undefined \def \bsertitle#1{#1}\fi
\ifx \bsnm \undefined \def \bsnm#1{#1}\fi
\ifx \bsuffix \undefined \def \bsuffix#1{#1}\fi
\ifx \bparticle \undefined \def \bparticle#1{#1}\fi
\ifx \barticle \undefined \def \barticle#1{#1}\fi
\bibcommenthead
\ifx \bconfdate \undefined \def \bconfdate #1{#1}\fi
\ifx \botherref \undefined \def \botherref #1{#1}\fi
\ifx \url \undefined \def \url#1{\textsf{#1}}\fi
\ifx \bchapter \undefined \def \bchapter#1{#1}\fi
\ifx \bbook \undefined \def \bbook#1{#1}\fi
\ifx \bcomment \undefined \def \bcomment#1{#1}\fi
\ifx \oauthor \undefined \def \oauthor#1{#1}\fi
\ifx \citeauthoryear \undefined \def \citeauthoryear#1{#1}\fi
\ifx \endbibitem  \undefined \def \endbibitem {}\fi
\ifx \bconflocation  \undefined \def \bconflocation#1{#1}\fi
\ifx \arxivurl  \undefined \def \arxivurl#1{\textsf{#1}}\fi
\csname PreBibitemsHook\endcsname

\bibitem[\protect\citeauthoryear{Jin et~al.}{2022}]{LSTM_SSSS}
\begin{barticle}
\bauthor{\bsnm{Jin}, \binits{Y.}},
\bauthor{\bsnm{Yu}, \binits{Y.}},
\bauthor{\bsnm{Chen}, \binits{C.}},
\bauthor{\bsnm{Zhao}, \binits{Z.}},
\bauthor{\bsnm{Heng}, \binits{P.-A.}},
\bauthor{\bsnm{Stoyanov}, \binits{D.}}:
\batitle{Exploring intra- and inter-video relation for surgical semantic scene segmentation}.
\bjtitle{IEEE Transactions on Medical Imaging}
\bvolume{41}(\bissue{11}),
\bfpage{2991}--\blpage{3002}
(\byear{2022})
\doiurl{10.1109/TMI.2022.3177077}
\end{barticle}
\endbibitem

\bibitem[\protect\citeauthoryear{Mohammed et~al.}{2019}]{StreoScenNet}
\begin{bchapter}
\bauthor{\bsnm{Mohammed}, \binits{A.}},
\bauthor{\bsnm{Yildirim}, \binits{S.}},
\bauthor{\bsnm{Farup}, \binits{I.}},
\bauthor{\bsnm{Pedersen}, \binits{M.}},
\bauthor{\bsnm{Hovde}, \binits{{\O}.}}:
\bctitle{{StreoScenNet: surgical stereo robotic scene segmentation}}.
In: \bbtitle{Medical Imaging 2019: Image-Guided Procedures, Robotic Interventions, and Modeling},
vol. \bseriesno{10951},
p. \bfpage{109510}.
\bpublisher{SPIE},
\blocation{~}
(\byear{2019}).
\doiurl{10.1117/12.2512518} .
\bcomment{International Society for Optics and Photonics}
\end{bchapter}
\endbibitem

\bibitem[\protect\citeauthoryear{Yoon et~al.}{2022}]{SimSSS}
\begin{bchapter}
\bauthor{\bsnm{Yoon}, \binits{J.}},
\bauthor{\bsnm{Hong}, \binits{S.}},
\bauthor{\bsnm{Hong}, \binits{S.}},
\bauthor{\bsnm{Lee}, \binits{J.}},
\bauthor{\bsnm{Shin}, \binits{S.}},
\bauthor{\bsnm{Park}, \binits{B.}},
\bauthor{\bsnm{Sung}, \binits{N.}},
\bauthor{\bsnm{Yu}, \binits{H.}},
\bauthor{\bsnm{Kim}, \binits{S.}},
\bauthor{\bsnm{Park}, \binits{S.}},
\bauthor{\bsnm{Hyung}, \binits{W.J.}},
\bauthor{\bsnm{Choi}, \binits{M.-K.}}:
\bctitle{Surgical scene segmentation using semantic image synthesis with a virtual surgery environment}.
In: \bbtitle{Medical Image Computing and Computer Assisted Intervention -- MICCAI 2022},
pp. \bfpage{551}--\blpage{561}.
\bpublisher{Springer},
\blocation{Cham}
(\byear{2022})
\end{bchapter}
\endbibitem

\bibitem[\protect\citeauthoryear{Fuentes-Hurtado et~al.}{2019}]{easylabels}
\begin{barticle}
\bauthor{\bsnm{Fuentes-Hurtado}, \binits{F.}},
\bauthor{\bsnm{Kadkhodamohammadi}, \binits{A.}},
\bauthor{\bsnm{Flouty}, \binits{E.}},
\bauthor{\bsnm{Barbarisi}, \binits{S.}},
\bauthor{\bsnm{Luengo}, \binits{I.}},
\bauthor{\bsnm{Stoyanov}, \binits{D.}}:
\batitle{Easylabels: weak labels for scene segmentation in laparoscopic videos}.
\bjtitle{International journal of computer assisted radiology and surgery}
\bvolume{14}(\bissue{7}),
\bfpage{1247}--\blpage{1257}
(\byear{2019})
\end{barticle}
\endbibitem

\bibitem[\protect\citeauthoryear{Allan et~al.}{2020}]{RSS}
\begin{botherref}
\oauthor{\bsnm{Allan}, \binits{M.}},
\oauthor{\bsnm{Kondo}, \binits{S.}},
\oauthor{\bsnm{Bodenstedt}, \binits{S.}},
\oauthor{\bsnm{Leger}, \binits{S.}},
\oauthor{\bsnm{Kadkhodamohammadi}, \binits{R.}},
\oauthor{\bsnm{Luengo}, \binits{I.}},
\oauthor{\bsnm{Fuentes}, \binits{F.}},
\oauthor{\bsnm{Flouty}, \binits{E.}},
\oauthor{\bsnm{Mohammed}, \binits{A.}},
\oauthor{\bsnm{Pedersen}, \binits{M.}},
\oauthor{\bsnm{Kori}, \binits{A.}},
\oauthor{\bsnm{Alex}, \binits{V.}},
\oauthor{\bsnm{Krishnamurthi}, \binits{G.}},
\oauthor{\bsnm{Rauber}, \binits{D.}},
\oauthor{\bsnm{Mendel}, \binits{R.}},
\oauthor{\bsnm{Palm}, \binits{C.}},
\oauthor{\bsnm{Bano}, \binits{S.}},
\oauthor{\bsnm{Saibro}, \binits{G.}},
\oauthor{\bsnm{Shih}, \binits{C.-S.}},
\oauthor{\bsnm{Chiang}, \binits{H.-A.}},
\oauthor{\bsnm{Zhuang}, \binits{J.}},
\oauthor{\bsnm{Yang}, \binits{J.}},
\oauthor{\bsnm{Iglovikov}, \binits{V.}},
\oauthor{\bsnm{Dobrenkii}, \binits{A.}},
\oauthor{\bsnm{Reddiboina}, \binits{M.}},
\oauthor{\bsnm{Reddy}, \binits{A.}},
\oauthor{\bsnm{Liu}, \binits{X.}},
\oauthor{\bsnm{Gao}, \binits{C.}},
\oauthor{\bsnm{Unberath}, \binits{M.}},
\oauthor{\bsnm{Kim}, \binits{M.}},
\oauthor{\bsnm{Kim}, \binits{C.}},
\oauthor{\bsnm{Kim}, \binits{C.}},
\oauthor{\bsnm{Kim}, \binits{H.}},
\oauthor{\bsnm{Lee}, \binits{G.}},
\oauthor{\bsnm{Ullah}, \binits{I.}},
\oauthor{\bsnm{Luna}, \binits{M.}},
\oauthor{\bsnm{Park}, \binits{S.H.}},
\oauthor{\bsnm{Azizian}, \binits{M.}},
\oauthor{\bsnm{Stoyanov}, \binits{D.}},
\oauthor{\bsnm{Maier-Hein}, \binits{L.}},
\oauthor{\bsnm{Speidel}, \binits{S.}}:
2018 Robotic Scene Segmentation Challenge.
arXiv
(2020).
\doiurl{10.48550/ARXIV.2001.11190}
\end{botherref}
\endbibitem

\bibitem[\protect\citeauthoryear{}{2021}]{HeiSurF}
\begin{botherref}
{HeiChole Surgical Workflow Analysis and Full Scene Segmentation (HeiSurF), EndoVis Subchallenge 2021}.
\url{https://www.synapse.org/\#!Synapse:syn25101790/wiki/608802}
Accessed 2022-11-14
\end{botherref}
\endbibitem

\bibitem[\protect\citeauthoryear{Maier-Hein et~al.}{2022}]{MaierHein22SDS}
\begin{barticle}
\bauthor{\bsnm{Maier-Hein}, \binits{L.}},
\bauthor{\bsnm{Eisenmann}, \binits{M.}},
\bauthor{\bsnm{Sarikaya}, \binits{D.}},
\bauthor{\bsnm{März}, \binits{K.}},
\bauthor{\bsnm{Collins}, \binits{T.}},
\bauthor{\bsnm{Malpani}, \binits{A.}},
\bauthor{\bsnm{Fallert}, \binits{J.}},
\bauthor{\bsnm{Feussner}, \binits{H.}},
\bauthor{\bsnm{Giannarou}, \binits{S.}},
\bauthor{\bsnm{Mascagni}, \binits{P.}},
\bauthor{\bsnm{Nakawala}, \binits{H.}},
\bauthor{\bsnm{Park}, \binits{A.}},
\bauthor{\bsnm{Pugh}, \binits{C.}},
\bauthor{\bsnm{Stoyanov}, \binits{D.}},
\bauthor{\bsnm{Vedula}, \binits{S.S.}},
\bauthor{\bsnm{Cleary}, \binits{K.}},
\bauthor{\bsnm{Fichtinger}, \binits{G.}},
\bauthor{\bsnm{Forestier}, \binits{G.}},
\bauthor{\bsnm{Gibaud}, \binits{B.}},
\bauthor{\bsnm{Grantcharov}, \binits{T.}},
\bauthor{\bsnm{Hashizume}, \binits{M.}},
\bauthor{\bsnm{Heckmann-Nötzel}, \binits{D.}},
\bauthor{\bsnm{Kenngott}, \binits{H.G.}},
\bauthor{\bsnm{Kikinis}, \binits{R.}},
\bauthor{\bsnm{Mündermann}, \binits{L.}},
\bauthor{\bsnm{Navab}, \binits{N.}},
\bauthor{\bsnm{Onogur}, \binits{S.}},
\bauthor{\bsnm{Roß}, \binits{T.}},
\bauthor{\bsnm{Sznitman}, \binits{R.}},
\bauthor{\bsnm{Taylor}, \binits{R.H.}},
\bauthor{\bsnm{Tizabi}, \binits{M.D.}},
\bauthor{\bsnm{Wagner}, \binits{M.}},
\bauthor{\bsnm{Hager}, \binits{G.D.}},
\bauthor{\bsnm{Neumuth}, \binits{T.}},
\bauthor{\bsnm{Padoy}, \binits{N.}},
\bauthor{\bsnm{Collins}, \binits{J.}},
\bauthor{\bsnm{Gockel}, \binits{I.}},
\bauthor{\bsnm{Goedeke}, \binits{J.}},
\bauthor{\bsnm{Hashimoto}, \binits{D.A.}},
\bauthor{\bsnm{Joyeux}, \binits{L.}},
\bauthor{\bsnm{Lam}, \binits{K.}},
\bauthor{\bsnm{Leff}, \binits{D.R.}},
\bauthor{\bsnm{Madani}, \binits{A.}},
\bauthor{\bsnm{Marcus}, \binits{H.J.}},
\bauthor{\bsnm{Meireles}, \binits{O.}},
\bauthor{\bsnm{Seitel}, \binits{A.}},
\bauthor{\bsnm{Teber}, \binits{D.}},
\bauthor{\bsnm{Ückert}, \binits{F.}},
\bauthor{\bsnm{Müller-Stich}, \binits{B.P.}},
\bauthor{\bsnm{Jannin}, \binits{P.}},
\bauthor{\bsnm{Speidel}, \binits{S.}}:
\batitle{Surgical data science – from concepts toward clinical translation}.
\bjtitle{Medical Image Analysis}
\bvolume{76},
\bfpage{102306}
(\byear{2022})
\doiurl{10.1016/j.media.2021.102306}
\end{barticle}
\endbibitem

\bibitem[\protect\citeauthoryear{Carstens et~al.}{2023}]{DSAD}
\begin{barticle}
\bauthor{\bsnm{Carstens}, \binits{M.}},
\bauthor{\bsnm{Rinner}, \binits{F.M.}},
\bauthor{\bsnm{Bodenstedt}, \binits{S.}},
\bauthor{\bsnm{Jenke}, \binits{A.C.}},
\bauthor{\bsnm{Weitz}, \binits{J.}},
\bauthor{\bsnm{Distler}, \binits{M.}},
\bauthor{\bsnm{Speidel}, \binits{S.}},
\bauthor{\bsnm{Kolbinger}, \binits{F.R.}}:
\batitle{The dresden surgical anatomy dataset for abdominal organ segmentation in surgical data science}.
\bjtitle{Scientific Data}
\bvolume{10}(\bissue{1}),
\bfpage{1}--\blpage{8}
(\byear{2023})
\doiurl{10.1038/s41597-022-01719-2}
\end{barticle}
\endbibitem

\bibitem[\protect\citeauthoryear{Shi et~al.}{2021}]{shi2021}
\begin{barticle}
\bauthor{\bsnm{Shi}, \binits{G.}},
\bauthor{\bsnm{Xiao}, \binits{L.}},
\bauthor{\bsnm{Chen}, \binits{Y.}},
\bauthor{\bsnm{Zhou}, \binits{S.K.}}:
\batitle{Marginal loss and exclusion loss for partially supervised multi-organ segmentation}.
\bjtitle{Medical Image Analysis}
\bvolume{70},
\bfpage{101979}
(\byear{2021})
\doiurl{10.1016/j.media.2021.101979}
\end{barticle}
\endbibitem

\bibitem[\protect\citeauthoryear{Ulrich et~al.}{2023}]{MultiTalent}
\begin{bchapter}
\bauthor{\bsnm{Ulrich}, \binits{C.}},
\bauthor{\bsnm{Isensee}, \binits{F.}},
\bauthor{\bsnm{Wald}, \binits{T.}},
\bauthor{\bsnm{Zenk}, \binits{M.}},
\bauthor{\bsnm{Baumgartner}, \binits{M.}},
\bauthor{\bsnm{Maier-Hein}, \binits{K.H.}}:
\bctitle{Multitalent: A multi-dataset approach to medical image segmentation}.
In: \bbtitle{Medical Image Computing and Computer Assisted Intervention -- MICCAI 2023},
pp. \bfpage{648}--\blpage{658}.
\bpublisher{Springer},
\blocation{Cham}
(\byear{2023})
\end{bchapter}
\endbibitem

\bibitem[\protect\citeauthoryear{Dmitriev and Kaufman}{2019}]{dmitriev2019learning}
\begin{bchapter}
\bauthor{\bsnm{Dmitriev}, \binits{K.}},
\bauthor{\bsnm{Kaufman}, \binits{A.E.}}:
\bctitle{Learning multi-class segmentations from single-class datasets}.
In: \bbtitle{Proceedings of the IEEE/CVF Conference on Computer Vision and Pattern Recognition},
pp. \bfpage{9501}--\blpage{9511}
(\byear{2019})
\end{bchapter}
\endbibitem

\bibitem[\protect\citeauthoryear{Yan et~al.}{2020}]{yan2020learning}
\begin{barticle}
\bauthor{\bsnm{Yan}, \binits{K.}},
\bauthor{\bsnm{Cai}, \binits{J.}},
\bauthor{\bsnm{Zheng}, \binits{Y.}},
\bauthor{\bsnm{Harrison}, \binits{A.P.}},
\bauthor{\bsnm{Jin}, \binits{D.}},
\bauthor{\bsnm{Tang}, \binits{Y.}},
\bauthor{\bsnm{Tang}, \binits{Y.}},
\bauthor{\bsnm{Huang}, \binits{L.}},
\bauthor{\bsnm{Xiao}, \binits{J.}},
\bauthor{\bsnm{Lu}, \binits{L.}}:
\batitle{Learning from multiple datasets with heterogeneous and partial labels for universal lesion detection in ct}.
\bjtitle{IEEE Transactions on Medical Imaging}
\bvolume{40}(\bissue{10}),
\bfpage{2759}--\blpage{2770}
(\byear{2020})
\end{barticle}
\endbibitem

\bibitem[\protect\citeauthoryear{Dice}{1945}]{dice}
\begin{barticle}
\bauthor{\bsnm{Dice}, \binits{L.R.}}:
\batitle{Measures of the amount of ecologic association between species}.
\bjtitle{Ecology}
\bvolume{26}(\bissue{3}),
\bfpage{297}--\blpage{302}
(\byear{1945})
\doiurl{10.2307/1932409}
\end{barticle}
\endbibitem

\bibitem[\protect\citeauthoryear{Kolbinger et~al.}{2023}]{dsad_split}
\begin{barticle}
\bauthor{\bsnm{Kolbinger}, \binits{F.R.}},
\bauthor{\bsnm{Rinner}, \binits{F.M.}},
\bauthor{\bsnm{Jenke}, \binits{A.C.}},
\bauthor{\bsnm{Carstens}, \binits{M.}},
\bauthor{\bsnm{Krell}, \binits{S.}},
\bauthor{\bsnm{Leger}, \binits{S.}},
\bauthor{\bsnm{Distler}, \binits{M.}},
\bauthor{\bsnm{Weitz}, \binits{J.}},
\bauthor{\bsnm{Speidel}, \binits{S.}},
\bauthor{\bsnm{Bodenstedt}, \binits{S.}}:
\batitle{Anatomy segmentation in laparoscopic surgery: comparison of machine learning and human expertise--an experimental study}.
\bjtitle{International Journal of Surgery}
\bvolume{109}(\bissue{10}),
\bfpage{2962}--\blpage{2974}
(\byear{2023})
\doiurl{10.1097/JS9.0000000000000595}
\end{barticle}
\endbibitem

\bibitem[\protect\citeauthoryear{Chen et~al.}{2017}]{deeplabv3}
\begin{botherref}
\oauthor{\bsnm{Chen}, \binits{L.-C.}},
\oauthor{\bsnm{Papandreou}, \binits{G.}},
\oauthor{\bsnm{Schroff}, \binits{F.}},
\oauthor{\bsnm{Adam}, \binits{H.}}:
Rethinking Atrous Convolution for Semantic Image Segmentation.
arXiv
(2017).
\doiurl{10.48550/arXiv.1706.05587}
\end{botherref}
\endbibitem

\bibitem[\protect\citeauthoryear{Lin et~al.}{2014}]{COCO}
\begin{botherref}
\oauthor{\bsnm{Lin}, \binits{T.-Y.}},
\oauthor{\bsnm{Maire}, \binits{M.}},
\oauthor{\bsnm{Belongie}, \binits{S.}},
\oauthor{\bsnm{Bourdev}, \binits{L.}},
\oauthor{\bsnm{Girshick}, \binits{R.}},
\oauthor{\bsnm{Hays}, \binits{J.}},
\oauthor{\bsnm{Perona}, \binits{P.}},
\oauthor{\bsnm{Ramanan}, \binits{D.}},
\oauthor{\bsnm{Zitnick}, \binits{C.L.}},
\oauthor{\bsnm{Dollár}, \binits{P.}}:
Microsoft COCO: Common Objects in Context.
arXiv
(2014).
\doiurl{10.48550/arxiv.1405.0312}
\end{botherref}
\endbibitem

\bibitem[\protect\citeauthoryear{Paszke et~al.}{2019}]{pytorch}
\begin{botherref}
\oauthor{\bsnm{Paszke}, \binits{A.}},
\oauthor{\bsnm{Gross}, \binits{S.}},
\oauthor{\bsnm{Massa}, \binits{F.}},
\oauthor{\bsnm{Lerer}, \binits{A.}},
\oauthor{\bsnm{Bradbury}, \binits{J.}},
\oauthor{\bsnm{Chanan}, \binits{G.}},
\oauthor{\bsnm{Killeen}, \binits{T.}},
\oauthor{\bsnm{Lin}, \binits{Z.}},
\oauthor{\bsnm{Gimelshein}, \binits{N.}},
\oauthor{\bsnm{Antiga}, \binits{L.}},
\oauthor{\bsnm{Desmaison}, \binits{A.}},
\oauthor{\bsnm{Köpf}, \binits{A.}},
\oauthor{\bsnm{Yang}, \binits{E.}},
\oauthor{\bsnm{DeVito}, \binits{Z.}},
\oauthor{\bsnm{Raison}, \binits{M.}},
\oauthor{\bsnm{Tejani}, \binits{A.}},
\oauthor{\bsnm{Chilamkurthy}, \binits{S.}},
\oauthor{\bsnm{Steiner}, \binits{B.}},
\oauthor{\bsnm{Fang}, \binits{L.}},
\oauthor{\bsnm{Bai}, \binits{J.}},
\oauthor{\bsnm{Chintala}, \binits{S.}}:
PyTorch: An Imperative Style, High-Performance Deep Learning Library.
arXiv
(2019).
\doiurl{10.48550/arxiv.1912.01703}
\end{botherref}
\endbibitem

\end{thebibliography}

\end{document}